\documentclass[letterpaper]{article}

\usepackage{times}
\usepackage{helvet}
\usepackage{courier}
\usepackage{arxiv}
\usepackage{verbatim}

\usepackage[utf8]{inputenc} 
\usepackage[T1]{fontenc}    
\usepackage{hyperref}       
\usepackage{url}            
\usepackage{booktabs}       

\usepackage{subfig}
\usepackage{nicefrac}       
\usepackage{microtype}      
\usepackage{graphicx}
\usepackage{natbib}
\usepackage{doi}

\usepackage{multirow, rotating}

\usepackage{amsmath, amsthm, amssymb}
\usepackage{geometry}
\usepackage{booktabs}
\usepackage{array}
\usepackage{xcolor}
\usepackage{enumitem}
\usepackage{tikz}
\usepackage{algorithm}
\usepackage{algpseudocode}
\usepackage{caption}
\usetikzlibrary{positioning, arrows.meta, calc, shapes.geometric}

\hypersetup{
  colorlinks = true,
  linkcolor  = blue!60!black,
  citecolor  = blue!60!black,
  urlcolor   = blue!60!black
}

\theoremstyle{plain}
\newtheorem{theorem}{Theorem}[section]

\theoremstyle{definition}
\newtheorem{definition}[theorem]{Definition}
\newtheorem{example}[theorem]{Example}

\theoremstyle{remark}
\newtheorem{remark}[theorem]{Remark}

\newcommand{\GTS}{\mathrm{G2S}}
\newcommand{\STG}{\mathrm{S2G}}
\newcommand{\wstar}{w^{*}}
\newcommand{\Sig}{\Sigma}

\title{Instruction set for the representation of graphs}

\author{ \href{https://orcid.org/0000-0001-8231-5687}{\includegraphics[scale=0.06]{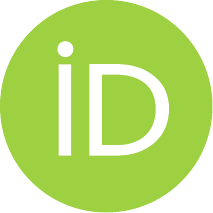}\hspace{1mm}Ezequiel L\'opez-Rubio}\thanks{Corresponding author. ITIS Software. Universidad de M\'alaga. C/ Arquitecto Francisco Peñalosa 18, 29010, Málaga, Spain} \\
	Department of Computer Languages and Computer Science\\
    University of M\'alaga\\
    Bulevar Louis Pasteur, 35\\
    29071 M\'alaga, Spain \\
	\texttt{ezeqlr@lcc.uma.es} \\
	\And
	\href{https://orcid.org/0000-0000-0000-0000}{\includegraphics[scale=0.06]{orcid.pdf}\hspace{1mm}Mario Pascual-Gonz\'alez} \\
    Department of Computer Languages and Computer Science\\
    University of M\'alaga\\
    Bulevar Louis Pasteur, 35\\
    29071 M\'alaga, Spain \\
	\texttt{mpascual@uma.es} \\
}

\renewcommand{\shorttitle}{Instruction set for the representation of graphs}

\hypersetup{
pdftitle={Instruction set for the representation of graphs},
pdfsubject={cs.AI, cs.CL},
pdfauthor={Ezequiel L\'opez-Rubio, Mario Pascual-Gonz\'alez},
pdfkeywords={graph representation,  adjacency matrix, instruction sequences, deep learning, language models, structural
patterns}
}

\begin{document}
\maketitle

\begin{abstract}
We present IsalGraph, a method for representing the structure of any
finite, simple graph as a compact string over a nine-character instruction
alphabet.  The encoding is
executed by a small virtual machine comprising a sparse graph, a circular
doubly-linked list (CDLL) of graph-node references, and two traversal
pointers.  Instructions either move a pointer through the CDLL or insert
a node or edge into the graph.  A key design property is that \emph{every}
string over $\Sigma$ decodes to a valid graph, with no invalid states
reachable.  A greedy \emph{GraphToString} algorithm encodes any connected
graph into a string in time polynomial in the number of nodes; an
exhaustive-backtracking variant produces a \emph{canonical string}
$\wstar_G$ by selecting the lexicographically smallest shortest string
across all starting nodes and all valid traversal orders.  We evaluate the representation on five
real-world graph benchmark datasets (IAM Letter LOW/MED/HIGH, LINUX, and
AIDS) and show that the Levenshtein
distance between IsalGraph strings correlates strongly with graph edit
distance (GED).  Together, these properties make IsalGraph
strings a compact, isomorphism-invariant, and language-model-compatible
sequential encoding of graph structure, with direct applications in
graph similarity search, graph generation, and graph-conditioned
language modelling.
\end{abstract}


\section{Introduction}

Graphs are among the most expressive data structures available to
scientists and engineers.  Molecular compounds, social networks,
knowledge bases, protein interaction networks, and circuit topologies can
all be modelled as collections of nodes connected by edges
\citep{zhou2020gnn, khoshraftar2024survey, ju2024comprehensive}.
A central challenge in contemporary computational graph processing is
\emph{representation}: how should the structure of a graph be encoded in
a form that supports efficient computation, generalisation, and downstream
learning?

The dominant answer is the \emph{adjacency matrix}.  Given a graph
$G = (V, E)$ on $N = |V|$ nodes, its adjacency matrix $M_G \in \{0,1\}^{N
\times N}$ records which pairs of nodes are connected.  The adjacency matrix is the foundation of spectral graph theory, algebraic graph algorithms, and virtually all existing deep learning approaches to graphs
\citep{kipf2017gcn, hamilton2017graphsage, velickovic2018gat}.  Its
limitations, however, are substantial: it occupies $O(N^2)$ space regardless of graph sparsity. Furthermore, it is inherently two-dimensional and
therefore not directly consumable by sequential models such as recurrent
networks or transformers. Last but not least, it breaks permutation equivariance because its meaning depends on the arbitrary ordering assigned to the nodes.

A possible alternative line of research seeks to encode graphs as \emph{sequences}
that can be fed to sequence models.  This is particularly
appealing in the current era of large language models, which have
demonstrated remarkable capacity to process, generate, and reason over
sequential data \citep{vaswani2017attention, devlin2019bert}.  The
challenge is to design a sequential encoding that is: (i) \emph{compact},
using much less than $O(N^2)$ symbols for sparse graphs; (ii) \emph{reversible}, so that
the original graph structure can be recovered exactly from the string;
(iii) \emph{structure-preserving}, so that similar graphs yield similar
strings; and (iv) \emph{canonicalisable}, admitting a unique representative
string per isomorphism class.

This paper presents \textbf{IsalGraph}, a novel methodology for sequential graph representation
satisfying all four desiderata.  The encoding is defined by a small
virtual machine comprising a sparse graph, a circular doubly linked list
(CDLL) of graph nodes, and two traversal pointers.  Nine instructions
move the pointers through the CDLL or insert nodes and edges into the
graph.  Executing any string in the instruction language decodes it into a
graph; conversely, a greedy algorithm encodes any connected graph into a
string. It must be highlighted that all strings over the defined alphabet are valid, i.e. they decode to a graph. A canonical string is obtained by minimising string length over
all starting nodes and all valid traversal orders, producing a complete
graph invariant with formal correctness guarantees. Our previous work \citep{lopezrubio2025isalgraph} is substantially different from IsalGraph because the older approach requires a fixed ordering of the nodes and does not employ a circular doubly linked list of nodes.






The structure of this paper is as follows. Section \ref{sec:method}
presents the graph representation methodology. After that, Section
\ref{sec:Computational-experiments} reports the results of an exploratory
computational experiment. Finally, Section \ref{sec:Conclusion} deals
with the conclusions.

\theoremstyle{plain}

\theoremstyle{definition}

\theoremstyle{remark}

\newtheorem{conjecture}[theorem]{Conjecture}

\newcommand{\IsalGraph}{\textsc{IsalGraph}}

\newcommand{\val}{\mathrm{val}}
\newcommand{\cdll}{\mathcal{L}}
\newcommand{\ptr}[1]{\pi_{#1}}
\newcommand{\N}{\mathbb{N}}

\algrenewcommand\algorithmiccomment[1]{\hfill\textcolor{gray!70!black}{\(\triangleright\) #1}}
\newcommand{\LineComment}[1]{\State \textcolor{gray!70!black}{\(\triangleright\) #1}}

\section{Methodology}
\label{sec:method}

This section presents the formal machinery of \IsalGraph{}.
Subsection~\ref{sec:instructions} defines the interpreter state and
instruction set.  Subsection~\ref{sec:s2g} describes the
\emph{StringToGraph} ($\STG$) algorithm.  Subsection~\ref{sec:g2s} presents
the \emph{GraphToString} ($\GTS$) algorithm.
Subsection~\ref{sec:conjecture} states the canonical-string invariance
conjecture.  Subsection~\ref{sec:topology} establishes the topological
relationship between the \IsalGraph{} string metric and graph edit
distance.

\subsection{Instruction Set and String Execution}
\label{sec:instructions}

\subsubsection{Interpreter State}

The \IsalGraph{} interpreter maintains three components simultaneously
during the execution of an instruction string.

\begin{definition}[Interpreter state]
\label{def:state}
An \emph{\IsalGraph{} interpreter state} is a triple
$\mathcal{S} = (G,\, \cdll,\, \pi)$ where:
\begin{itemize}
  \item $G = (V_G, E_G)$ is a finite, simple graph built incrementally,
        with nodes identified by contiguous non-negative integers
        $\{0, 1, \ldots, |V_G|-1\}$.
  \item $\cdll$ is an \emph{array-backed circular doubly-linked list}
        (CDLL) whose nodes carry graph-node indices as integer
        \emph{payloads}.  We write $\val_\cdll(\ell)$ for the payload of
        CDLL node $\ell$, and $\mathrm{next}(\ell)$,
        $\mathrm{prev}(\ell)$ for its successor and predecessor in the
        circular order.  The CDLL index space and the graph node index
        space are \emph{distinct}: a CDLL node $\ell$ is not the same
        object as graph node $\val_\cdll(\ell)$.
  \item $\pi = (\ptr{1}, \ptr{2})$ is a pair of \emph{pointers}, where
        each pointer is a CDLL node index.  $\ptr{1}$ is called the
        \emph{primary pointer} and $\ptr{2}$ the \emph{secondary
        pointer}.
\end{itemize}
\end{definition}

\paragraph{Initial state.}
Before any instruction is executed, the interpreter is placed in the
following \emph{initial state}:
\begin{enumerate}[label=(\roman*)]
  \item $G$ contains exactly one node (node~$0$) and no edges.
  \item $\cdll$ contains exactly one node whose payload is graph node~$0$.
  \item Both pointers $\ptr{1}$ and $\ptr{2}$ point to this single CDLL
        node.
\end{enumerate}

\subsubsection{The Instruction Alphabet}

\IsalGraph{} strings are drawn from the nine-character alphabet
\[
  \Sig \;=\; \{N,\, n,\, P,\, p,\, V,\, v,\, C,\, c,\, W\}.
\]
The semantics of each instruction are defined in
Table~\ref{tab:instructions} and expanded below.

\begin{table}[ht]
\caption{The \IsalGraph{} instruction set.  $\val_\cdll(\ell)$ denotes
         the graph-node index stored as payload of CDLL node $\ell$.
         The $N/n$ ($P/p$) instructions traverse the CDLL in the
         forward (backward) circular direction.  The $V/v$ instructions
         always create edges \emph{from} the pointer node \emph{to} the
         new node; the pointer itself does \emph{not} move.
         Instructions $C$ and $c$ differ only for directed graphs.}
\centering
\small
\renewcommand{\arraystretch}{1.4}
\begin{tabular}{p{1.1cm} p{4.2cm} p{7.2cm}}
\toprule
\textbf{Instr.} & \textbf{Type} & \textbf{Effect on state
$(G, \cdll, \ptr{1}, \ptr{2})$} \\
\midrule
$N$ & Primary move (forward)
    & $\ptr{1} \leftarrow \mathrm{next}_\cdll(\ptr{1})$ \\
$P$ & Primary move (backward)
    & $\ptr{1} \leftarrow \mathrm{prev}_\cdll(\ptr{1})$ \\
$n$ & Secondary move (forward)
    & $\ptr{2} \leftarrow \mathrm{next}_\cdll(\ptr{2})$ \\
$p$ & Secondary move (backward)
    & $\ptr{2} \leftarrow \mathrm{prev}_\cdll(\ptr{2})$ \\
$V$ & Node insertion via primary
    & Add new node $u$ to $G$; add edge $(\val_\cdll(\ptr{1}),\, u)$ to
      $G$; insert $u$ into $\cdll$ immediately after $\ptr{1}$. \\
$v$ & Node insertion via secondary
    & Add new node $u$ to $G$; add edge $(\val_\cdll(\ptr{2}),\, u)$ to
      $G$; insert $u$ into $\cdll$ immediately after $\ptr{2}$. \\
$C$ & Edge insertion (primary $\to$ secondary)
    & Add edge $(\val_\cdll(\ptr{1}),\, \val_\cdll(\ptr{2}))$ to $G$.
      For undirected graphs, the reverse edge is also added. \\
$c$ & Edge insertion (secondary $\to$ primary)
    & Add edge $(\val_\cdll(\ptr{2}),\, \val_\cdll(\ptr{1}))$ to $G$.
      Equivalent to $C$ for undirected graphs. \\
$W$ & No-op
    & State is unchanged. \\
\bottomrule
\end{tabular}

\label{tab:instructions}
\end{table}

\paragraph{Critical semantic note.}
In the $V$ and $v$ instructions, the new CDLL node for $u$ is inserted
\emph{after} the pointer's current CDLL node, but the pointer itself does
not advance to the new node.  This means that after a $V$ instruction,
$\ptr{1}$ still references the same CDLL node as before the instruction
was executed.

\paragraph{Every string is valid.}
A key design property of the \IsalGraph{} alphabet is that \emph{every}
string $w \in \Sig^*$ decodes to a valid finite simple graph.  No
instruction can produce an undefined or inconsistent state: pointer
movements wrap around the circular CDLL, and node- and edge-insertion
instructions always have a well-defined, deterministic effect.

\subsubsection{The StringToGraph Algorithm\label{sec:s2g}}

The $\STG$ algorithm (Algorithm~\ref{alg:s2g}) executes an \IsalGraph{}
string instruction by instruction, starting from the initial state and
returning the resulting graph.

\begin{algorithm}[ht]
\caption{$\STG(w,\, \mathit{directed})$: StringToGraph}
\label{alg:s2g}
\begin{algorithmic}[1]
\Require String $w \in \Sig^*$; Boolean $\mathit{directed}$
\Ensure  Graph $G$ such that $\STG(w) = G$

\LineComment{Initialise interpreter state}
\State $G \leftarrow$ new graph with one node $u_0 = 0$, directed $=
       \mathit{directed}$
\State $\cdll \leftarrow$ new CDLL; $\ell_0 \leftarrow
       \cdll.\mathrm{insert\_after}(\varnothing,\; u_0)$
\State $\ptr{1} \leftarrow \ell_0$;\quad $\ptr{2} \leftarrow \ell_0$

\LineComment{Execute each instruction in turn}
\For{each character $\sigma$ in $w$}
  \If{$\sigma = N$}
    \State $\ptr{1} \leftarrow \cdll.\mathrm{next}(\ptr{1})$
  \ElsIf{$\sigma = P$}
    \State $\ptr{1} \leftarrow \cdll.\mathrm{prev}(\ptr{1})$
  \ElsIf{$\sigma = n$}
    \State $\ptr{2} \leftarrow \cdll.\mathrm{next}(\ptr{2})$
  \ElsIf{$\sigma = p$}
    \State $\ptr{2} \leftarrow \cdll.\mathrm{prev}(\ptr{2})$
  \ElsIf{$\sigma = V$}
    \State $u \leftarrow G.\mathrm{add\_node}()$
    \State $G.\mathrm{add\_edge}\!\bigl(\val_\cdll(\ptr{1}),\; u\bigr)$
    \State $\cdll.\mathrm{insert\_after}(\ptr{1},\; u)$
    \Comment{pointer $\ptr{1}$ does not move}
  \ElsIf{$\sigma = v$}
    \State $u \leftarrow G.\mathrm{add\_node}()$
    \State $G.\mathrm{add\_edge}\!\bigl(\val_\cdll(\ptr{2}),\; u\bigr)$
    \State $\cdll.\mathrm{insert\_after}(\ptr{2},\; u)$
    \Comment{pointer $\ptr{2}$ does not move}
  \ElsIf{$\sigma = C$}
    \State $G.\mathrm{add\_edge}\!\bigl(\val_\cdll(\ptr{1}),\;
           \val_\cdll(\ptr{2})\bigr)$
  \ElsIf{$\sigma = c$}
    \State $G.\mathrm{add\_edge}\!\bigl(\val_\cdll(\ptr{2}),\;
           \val_\cdll(\ptr{1})\bigr)$
  \ElsIf{$\sigma = W$}
    \State \textbf{skip} \Comment{no-op}
  \EndIf
\EndFor
\State \Return $G$
\end{algorithmic}
\end{algorithm}

\begin{remark}
For undirected graphs, \texttt{add\_edge$(u, v)$} inserts both $(u,v)$
and $(v,u)$ into the adjacency structure, so instructions $C$ and $c$
have identical effect.  For directed graphs they differ by edge
direction.
\end{remark}

\begin{example}[Decoding \texttt{VvNV}]
\label{ex:decode}
We trace $\STG(\texttt{VvNV},\; \mathit{directed}=\mathit{false})$:
\begin{enumerate}[label=(\arabic*)]
  \item \emph{Init}: $G = (\{0\},\, \varnothing)$; CDLL $= [0]$;
        $\ptr{1}=\ptr{2}=\ell_0$ (payload~$0$).
  \item \texttt{V}: add node $1$; add edge $(0,1)$; CDLL $= [0,1]$;
        $\ptr{1}$ still on $\ell_0$.
  \item \texttt{v}: add node $2$; add edge $(0,2)$; CDLL $= [0,1,2]$
        (inserted after $\ptr{2}=\ell_0$, so after~$0$ but before~$1$
        in circular order --- actually $[0,2,1]$); $\ptr{2}$ still on $\ell_0$.
  \item \texttt{N}: $\ptr{1} \leftarrow \mathrm{next}(\ell_0)$ (node~$2$
        in current circular order~$[0,2,1]$).
  \item \texttt{V}: add node $3$; add edge $(2,3)$; CDLL $= [0,2,3,1]$.
\end{enumerate}
\end{example}

\subsection{Graph-to-String Conversion}
\label{sec:g2s}

The $\GTS$ algorithm is the inverse of $\STG$: given a graph $G$ and a
starting node $v_0 \in V(G)$, it produces an \IsalGraph{} string $w$
such that $\STG(w) \cong G$.  The algorithm is a \emph{greedy search}
that at each step finds the cheapest pointer displacement (in terms of
number of pointer-move instructions emitted) that enables a useful
structural operation.

\subsubsection{Pair Generation and Cost Ordering}

The search space at each step is the set of integer displacement pairs
$(a, b) \in \{-M, \ldots, M\}^2$, where $M$ is the current node count
and $a$, $b$ are the number of steps to move the primary and secondary
pointers respectively (positive = forward, negative = backward).  The
\emph{cost} of a pair is its total pointer-movement count $|a| + |b|$,
which equals the number of $N$/$P$/$n$/$p$ instructions that will be
emitted.

\begin{definition}[Sorted displacement pairs]
\label{def:pairs}
For a positive integer $M$, let
\[
  \mathcal{P}(M) \;=\; \Bigl\{(a,b) \;\Big|\; a, b \in [-M, M]\Bigr\}
\]
sorted in increasing order of $(|a|+|b|,\; |a|,\; a,\; b)$
lexicographically.  The primary sort key $|a|+|b|$ minimises total
pointer movement; secondary keys break ties deterministically.
\end{definition}

\subsubsection{Algorithm Description}

The algorithm (Algorithm~\ref{alg:g2s}) maintains an output graph
$G_{\mathrm{out}}$ and two node-index mappings: $\iota$ (input-to-output)
and $\iota^{-1}$ (output-to-input).  These mappings are necessary because
$G_{\mathrm{out}}$ is built incrementally with its own node numbering,
which may differ from the input graph's numbering.

At each iteration, the algorithm enumerates pairs $(a,b) \in
\mathcal{P}(M)$ in order and attempts four operations in priority order:
\begin{description}
  \item[$V$ (node via primary)] The tentative primary position
        $\tilde{\pi}_1$ corresponds to a node in the input graph that
        has an \emph{unmapped} neighbour.  A new node is inserted.
  \item[$v$ (node via secondary)] Same, but using the tentative secondary
        position $\tilde{\pi}_2$.
  \item[$C$ (edge, primary $\to$ secondary)] The tentative positions
        correspond to two nodes in the input graph that are adjacent but
        whose corresponding output-graph nodes are not yet connected.
  \item[$c$ (edge, secondary $\to$ primary)] Same as $C$ but reversed;
        meaningful only for directed graphs.
\end{description}
The first pair $(a,b)$ for which any of these operations is applicable is
committed: the pointer-move instructions are emitted ($|a|$ copies of $N$
or $P$; $|b|$ copies of $n$ or $p$), the structural instruction ($V$,
$v$, $C$, or $c$) is appended, the actual pointers are updated, and the
loop continues.  The algorithm terminates when all nodes and all edges of
$G$ have been inserted.

\begin{algorithm}[ht]
\caption{$\GTS(G,\, v_0)$: GraphToString (greedy)}
\label{alg:g2s}
\begin{algorithmic}[1]
\Require Connected graph $G=(V,E)$; starting node $v_0 \in V$
\Ensure  String $w \in \Sig^*$ with $\STG(w) \cong G$

\LineComment{Initialise state}
\State Verify all nodes are reachable from $v_0$
\State $G_{\mathrm{out}} \leftarrow$ empty graph; $u_0 \leftarrow
       G_{\mathrm{out}}.\mathrm{add\_node}()$
\State $\cdll \leftarrow$ new CDLL;\quad $\ell_0 \leftarrow
       \cdll.\mathrm{insert\_after}(\varnothing,\; u_0)$
\State $\ptr{1} \leftarrow \ell_0$;\quad $\ptr{2} \leftarrow \ell_0$
\State $\iota \leftarrow \{v_0 \mapsto u_0\}$;\quad
       $\iota^{-1} \leftarrow \{u_0 \mapsto v_0\}$
\State $n_{\mathrm{left}} \leftarrow |V|-1$;\quad
       $e_{\mathrm{left}} \leftarrow |E|$;\quad $w \leftarrow \varepsilon$

\LineComment{Main loop: continue until all nodes and edges are inserted}
\While{$n_{\mathrm{left}} > 0$ \textbf{ or } $e_{\mathrm{left}} > 0$}
  \State $M \leftarrow |V(G_{\mathrm{out}})|$
  \For{$(a, b)$ \textbf{in} $\mathcal{P}(M)$}
    \State $\tilde{\ell}_1 \leftarrow \mathrm{walk}(\cdll,\, \ptr{1},\, a)$;\quad
           $\tilde{v}_1 \leftarrow \iota^{-1}[\val_\cdll(\tilde{\ell}_1)]$
    \State $\tilde{\ell}_2 \leftarrow \mathrm{walk}(\cdll,\, \ptr{2},\, b)$;\quad
           $\tilde{v}_2 \leftarrow \iota^{-1}[\val_\cdll(\tilde{\ell}_2)]$

    \If{$n_{\mathrm{left}}>0$ \textbf{ and } $\exists\, c \in N_G(\tilde{v}_1)$ with
        $c \notin \mathrm{dom}(\iota)$} \Comment{$V$: node via primary}
      \State $u \leftarrow G_{\mathrm{out}}.\mathrm{add\_node}()$;\;
             $\iota[c]\leftarrow u$;\; $\iota^{-1}[u]\leftarrow c$
      \State $G_{\mathrm{out}}.\mathrm{add\_edge}(\val_\cdll(\tilde{\ell}_1),\; u)$;\;
             $\cdll.\mathrm{insert\_after}(\tilde{\ell}_1,\; u)$
      \State $w \mathrel{+}= \mathrm{moves}(a, \mathrm{primary}) + \texttt{V}$;\;
             $\ptr{1} \leftarrow \tilde{\ell}_1$
      \State $n_{\mathrm{left}} \mathrel{-}= 1$;\; $e_{\mathrm{left}} \mathrel{-}= 1$;\;
             \textbf{break}
    \ElsIf{$n_{\mathrm{left}}>0$ \textbf{ and } $\exists\, c \in N_G(\tilde{v}_2)$
           with $c \notin \mathrm{dom}(\iota)$} \Comment{$v$: node via secondary}
      \State $u \leftarrow G_{\mathrm{out}}.\mathrm{add\_node}()$;\;
             $\iota[c]\leftarrow u$;\; $\iota^{-1}[u]\leftarrow c$
      \State $G_{\mathrm{out}}.\mathrm{add\_edge}(\val_\cdll(\tilde{\ell}_2),\; u)$;\;
             $\cdll.\mathrm{insert\_after}(\tilde{\ell}_2,\; u)$
      \State $w \mathrel{+}= \mathrm{moves}(b, \mathrm{secondary}) + \texttt{v}$;\;
             $\ptr{2} \leftarrow \tilde{\ell}_2$
      \State $n_{\mathrm{left}} \mathrel{-}= 1$;\; $e_{\mathrm{left}} \mathrel{-}= 1$;\;
             \textbf{break}
    \ElsIf{$(\tilde{v}_2, \tilde{v}_1) \in E$ \textbf{ and }
           $(\val_\cdll(\tilde{\ell}_2), \val_\cdll(\tilde{\ell}_1))
           \notin E(G_{\mathrm{out}})$} \Comment{$C$}
      \State $G_{\mathrm{out}}.\mathrm{add\_edge}(\val_\cdll(\tilde{\ell}_1),\;
             \val_\cdll(\tilde{\ell}_2))$
      \State $w \mathrel{+}= \mathrm{moves}(a,\mathrm{pri}) +
             \mathrm{moves}(b,\mathrm{sec}) + \texttt{C}$
      \State $\ptr{1} \leftarrow \tilde{\ell}_1$;\; $\ptr{2} \leftarrow \tilde{\ell}_2$;\;
             $e_{\mathrm{left}} \mathrel{-}= 1$;\; \textbf{break}
    \ElsIf{$G$ directed \textbf{ and } $(\tilde{v}_1, \tilde{v}_2) \in E$
           \textbf{ and } $(\val_\cdll(\tilde{\ell}_1), \val_\cdll(\tilde{\ell}_2))
           \notin E(G_{\mathrm{out}})$} \Comment{$c$}
      \State $G_{\mathrm{out}}.\mathrm{add\_edge}(\val_\cdll(\tilde{\ell}_2),\;
             \val_\cdll(\tilde{\ell}_1))$
      \State $w \mathrel{+}= \mathrm{moves}(a,\mathrm{pri}) +
             \mathrm{moves}(b,\mathrm{sec}) + \texttt{c}$
      \State $\ptr{1} \leftarrow \tilde{\ell}_1$;\; $\ptr{2} \leftarrow \tilde{\ell}_2$;\;
             $e_{\mathrm{left}} \mathrel{-}= 1$;\; \textbf{break}
    \EndIf
  \EndFor
\EndWhile
\State \Return $w$
\end{algorithmic}
\end{algorithm}

\noindent Here $\mathrm{walk}(\cdll, \ell, a)$ returns the CDLL node
reached by taking $|a|$ steps forward (if $a>0$) or backward (if $a<0$)
from $\ell$.  The helper $\mathrm{moves}(a, \mathrm{primary})$ emits
$a$ copies of $N$ (if $a \geq 0$) or $|a|$ copies of $P$ (if $a < 0$),
and analogously for the secondary pointer with $n/p$.

\begin{remark}[Reachability precondition]
\label{rem:reachability}
For directed graphs, the $V$ and $v$ instructions always create edges of
the form $(\mathrm{existing\_node} \to \mathrm{new\_node})$.  Consequently,
$\GTS$ can only encode nodes that are reachable from $v_0$ via directed
outgoing edges.  The algorithm raises an error if any node is unreachable
from the chosen starting node.
\end{remark}

\begin{remark}[String length decomposition]
\label{rem:length}
For a graph $G$ with $N$ nodes and $M$ edges, the length of any
\IsalGraph{} string encoding $G$ satisfies:
\[
  |w| \;=\; \underbrace{(N-1)}_{\text{one }V/v\text{ per non-root node}}
           \;+\; \underbrace{M - (N-1)}_{\text{one }C/c/V/v\text{ per extra edge}}
           \;+\; \underbrace{\sum_k (|a_k| + |b_k|)}_{\text{pointer moves}}.
\]
The first two terms are fixed by $G$; only the total pointer-movement
cost depends on the traversal order.  Minimising $|w|$ therefore reduces
to minimising total pointer travel.
\end{remark}

\subsection{Conjectured Properties}
\label{sec:conjecture}

The greedy $\GTS$ algorithm is not \emph{label-blind} in its base form:
its neighbour iteration order (over sets) depends
on the order that the nodes are extracted from the set, so two isomorphic graphs
with different node numberings may yield different strings from the greedy
algorithm.  To recover a labelling-independent encoding, we define the
\emph{canonical string} via exhaustive backtracking.

\begin{definition}[Canonical string]
\label{def:canonical}
Let $\mathcal{W}(G)$ denote the set of all \IsalGraph{} strings
producible by the exhaustive-backtracking variant of $\GTS$ (which
explores all valid neighbour choices at every $V/v$ branch point) over
all starting nodes $v \in V(G)$.  The \emph{canonical string} of $G$ is
\[
  \wstar_G \;=\; \mathrm{lexmin}
    \Bigl\{\,
      w \in \mathcal{W}(G)
      \;\Big|\;
      |w| = \min_{w' \in \mathcal{W}(G)} |w'|
    \Bigr\}.
\]
That is, among all shortest strings in $\mathcal{W}(G)$, select the
lexicographically smallest under the total order
$C < N < P < V < W < c < n < p < v$ on $\Sig$.
\end{definition}

This construction motivates the following conjecture, which we state
here as our primary theoretical claim and support empirically in
Section~\ref{sec:experiments}.

\begin{conjecture}[Canonical string as complete graph invariant]
\label{conj:invariant}
Let $G$ and $H$ be finite, simple graphs.  Then
\[
  G \cong H \;\iff\; \wstar_G = \wstar_H.
\]
\end{conjecture}

The forward direction ($G \cong H \Rightarrow \wstar_G = \wstar_H$) would
follow from the label-blindness of the exhaustive canonical search: an
isomorphism $\phi: V(G) \to V(H)$ bijects the set of valid traversals of
$G$ from $v$ onto the valid traversals of $H$ from $\phi(v)$, so the two
graphs generate identical sets of strings and hence the same canonical
minimum.  The backward direction ($\wstar_G = \wstar_H \Rightarrow G
\cong H$) would follow from round-trip correctness: if both $G$ and $H$
produce the same canonical string $w$, then
$G \cong \STG(w) \cong H$ by transitivity.

A complete proof requires establishing, rigorously, that the
exhaustive-backtracking algorithm is indeed label-blind, i.e.\ that its
output depends only on the abstract adjacency structure of the input and
not on the integer identifiers assigned to nodes.  We leave this
verification as future work and instead provide empirical support:
100\% invariance and discrimination rates on 71 isomorphic and
non-isomorphic graph pairs across nine graph families (see
Section~\ref{sec:experiments}).

\begin{remark}[Relation to graph isomorphism]
\label{rem:gi}
If Conjecture~\ref{conj:invariant} holds, then computing $\wstar_G$ is
at least as hard as graph isomorphism, since $\wstar_G = \wstar_H$ if
and only if $G \cong H$.  Graph isomorphism is known to lie in $\mathsf{NP}$
and is not known to be $\mathsf{NP}$-complete; it has quasi-polynomial-time
algorithms.  The exhaustive canonical search used to
compute $\wstar_G$ has complexity that grows super-polynomially with
$|V(G)|$ in the worst case.
\end{remark}

\subsection{Topological Structure}
\label{sec:topology}

A key property of a useful graph representation is \emph{metric
locality}: small structural changes to a graph should produce small
changes in its representation.  Conversely, structurally dissimilar
graphs should have representations that are far apart.  We formalise this
via a comparison between the \emph{Levenshtein distance} on \IsalGraph{}
strings and the standard \emph{Graph Edit Distance} (GED).

\subsubsection{The String Distance}

\begin{definition}[Levenshtein distance on \IsalGraph{} strings]
\label{def:levenshtein}
For two \IsalGraph{} strings $w_1, w_2 \in \Sig^*$, their
\emph{Levenshtein distance} is
\[
  d_{\mathrm{Lev}}(w_1, w_2)
  \;=\;
  \min\bigl\{
    k \;\big|\;
    w_1 \xrightarrow{\text{$k$ edits}} w_2
  \bigr\},
\]
where a single \emph{edit} is a character insertion, deletion, or
substitution.  This is computed in $O(|w_1| \cdot |w_2|)$ time via
standard dynamic programming.  Applied to canonical strings, we define
the \emph{\IsalGraph{} graph distance}:
\[
  d_{\IsalGraph}(G, H) \;=\; d_{\mathrm{Lev}}(\wstar_G,\, \wstar_H).
\]
\end{definition}

\subsubsection{Graph Edit Distance}

\begin{definition}[Graph Edit Distance~\citep{sanfeliu1983ged}]
\label{def:ged}
The \emph{Graph Edit Distance} $\mathrm{GED}(G, H)$ is the minimum
number of elementary edit operations (node insertion, node deletion, edge
insertion, edge deletion) needed to transform $G$ into a graph isomorphic
to $H$, under uniform unit costs.
\end{definition}

GED is a complete metric on the space of finite graphs up to isomorphism
(i.e.\ $\mathrm{GED}(G,H)=0 \iff G\cong H$), but it is
$\mathsf{NP}$-hard to compute even for simple unit-cost
functions.

\subsubsection{The Locality Property}

We state the locality relationship between $d_{\IsalGraph}$ and GED as a
claim. Let $G$ and $H$ be finite, simple, connected graphs.  Denote by
$k = \mathrm{GED}(G,H)$ their graph edit distance.  Then, empirically:
\begin{enumerate}[label=(\roman*)]
  \item \textbf{Monotonicity.} $d_{\IsalGraph}(G, H)$ is a
        non-decreasing function of $k$: adding more edit operations to
        $G$ produces a canonical string further from $\wstar_G$.
  \item \textbf{Strong correlation.} Over a broad sample of graph
        families, the Pearson correlation and the Spearman
        rank correlation between $d_{\IsalGraph}$ and
        GED are high.
  \item \textbf{Sensitivity.}  The mapping $k \mapsto d_{\IsalGraph}$ is
        monotonically increasing on average.
\end{enumerate}

The locality property distinguishes \IsalGraph{} from many other graph
representations.  For comparison, the Hamming distance between
(permuted) adjacency matrices does not satisfy locality because a
single node insertion changes $O(N)$ entries in the matrix.  The
\IsalGraph{} encoding instead reflects the \emph{instruction-level
cost} of re-encoding the modified graph, which is naturally bounded by
the number of changed edges plus the additional pointer moves required to
reach the new positions.

For the $N \times N$ binary adjacency matrix, a single edge insertion
changes exactly one entry (two for undirected graphs), giving a Hamming
distance of $1$ or $2$ per edge edit---asymptotically smaller than the
\IsalGraph{} bound.  However, the adjacency matrix does not admit a
meaningful string metric without first fixing a canonical node ordering,
which reintroduces the isomorphism problem.  The \IsalGraph{} distance
$d_{\IsalGraph}$ is isomorphism-invariant by construction, whereas the
Hamming distance on adjacency matrices is not.

\subsubsection{Implications for Graph Similarity Search}

The locality property has practical consequences.  First, it suggests
that $d_{\IsalGraph}$ can serve as a \emph{computationally efficient
proxy} for GED in similarity search: computing $d_{\mathrm{Lev}}$ takes
$O(|w_1| \cdot |w_2|)$ time, whereas exact GED requires exponential
time.  Second, the correlation is strong enough that
rankings produced by $d_{\IsalGraph}$ closely mirror GED rankings,
which is the property required for $k$-nearest-neighbour retrieval.
Third, because every \IsalGraph{} string decodes to a valid graph,
interpolation in the string space (e.g.\ via random or guided edit
paths) produces valid intermediate graphs, enabling gradient-free
graph optimisation via string-space random walks.

\section{Computational experiments\label{sec:Computational-experiments}}

\label{sec:experiments}

This section describes the datasets, evaluation protocol, and computational
infrastructure underlying the experiments.
Three objectives guide the experimental design:
(i)~quantifying the agreement between Levenshtein distance on \IsalGraph{} strings
and graph edit distance across real-world graph benchmarks;
(ii)~characterising the empirical time complexity of the encoding algorithms
on synthetic random graphs;
and (iii)~measuring the trade-off between encoding quality and computational cost
across three encoding strategies of increasing expense.

\subsection{Benchmark Datasets}
\label{sec:datasets}

The experiments require two kinds of benchmark data:
real-world graph collections with exact GED ground truth,
for evaluating how faithfully the Levenshtein distance
approximates structural dissimilarity;
and synthetic random graphs of controlled size,
for measuring how encoding time scales with the number of nodes.

\subsubsection{Real-World Graph Collections}

Five datasets from three application domains are used
for the correlation analysis and the speedup measurement.
In all cases, node and edge attributes are discarded;
the \IsalGraph{} encoding operates solely on graph topology.
Only connected graphs are retained, since the $\GTS$ algorithm
(Algorithm~\ref{alg:g2s}) requires a connected input.

\paragraph{IAM Letter (LOW, MED, HIGH).}
The IAM Letter dataset~\citep{riesen2008iam}
contains prototype graphs of 15 capital letters
of the Roman alphabet that consist exclusively of straight lines.
Nodes represent characteristic points of the letter strokes
(endpoints, corners, intersections), and edges connect consecutive
points along a stroke.
Three subsets correspond to increasing levels of positional noise
applied to node coordinates: \textsc{Low}, \textsc{Med}, and \textsc{High}.
After connectivity filtering, the subsets contain
1{,}180, 1{,}253, and 2{,}059 graphs, with mean edge counts
of 3.07, 3.17, and 4.56, respectively.
For these three subsets, exact GED is computed
via the A$^*$ algorithm of NetworkX~\citep{hagberg2008networkx}
with uniform unit costs:
node insertion and deletion cost~1,
edge insertion and deletion cost~1,
and node substitution cost~0
(all nodes are structurally identical after stripping coordinates).

\paragraph{LINUX.}
The LINUX dataset contains program flow graphs extracted from
subroutines of the Linux kernel,
originally collected by~\citet{bai2019simgnn}
and redistributed with isomorphism deduplication
by~\citet{jain2024graphedx}.
After filtering for connectivity and restricting to graphs
with at most 12 nodes, 89 graphs remain
(mean edge count: 8.35).
Precomputed exact GED matrices from the GraphEdX repository
are used directly;
these were obtained via A$^*$ search with topology-only costs
(zero cost for all node operations;
unit cost for edge insertion and deletion).

\paragraph{AIDS.}
The AIDS dataset contains molecular graphs from the
Developmental Therapeutics Program of the U.S.\ National Cancer Institute,
where nodes represent atoms and edges represent covalent bonds.
The topology-only variant distributed by~\citet{jain2024graphedx}
is used, in which all node labels (atom types) have been stripped
and GED is computed with the same topology-only cost function as LINUX.
After filtering, 769 connected graphs remain
(mean edge count: 10.70).

The five datasets cover structural densities
ranging from sparse (mean edges 3.07) to moderately dense (10.70),
and sample sizes from 89 to 2{,}059 graphs.
Table~\ref{tab:performance-summary} summarises the number of graphs,
the number of valid pairwise comparisons, and the mean edge count
for each dataset.

\subsubsection{Synthetic Graph Families}

The complexity characterisation requires graphs
of controlled size, independent of any particular application domain.
Random connected graphs are generated from two standard families:

\begin{itemize}[nosep]
  \item Barab\'{a}si--Albert (BA) preferential-attachment
        graphs~\citep{barabasi1999emergence}
        with attachment parameters $m \in \{1, 2\}$.
  \item Erd\H{o}s--R\'{e}nyi (ER) random
        graphs~\citep{erdos1959random}
        with edge probabilities $p \in \{0.3, 0.5\}$.
        When an ER graph is disconnected,
        only its largest connected component is retained.
\end{itemize}

For each family, graphs are generated at node counts
$n \in \{3, 4, \ldots, 50\}$ for the greedy methods
and $n \in \{3, 4, \ldots, 20\}$ for the canonical method,
with a per-instance timeout of 600 seconds.
Five independent instances are generated per
$(n, \text{family})$ combination.
These synthetic graphs are used exclusively
for the time-complexity analysis reported
in Figure~\ref{fig:empirical-complexity};
they do not participate in the correlation experiments.

\subsection{Evaluation Protocol}
\label{sec:protocol}

The evaluation is organised into four components:
a comparison of three encoding methods (Section~\ref{sec:enc-methods}),
a correlation analysis between Levenshtein and GED distances
on the real-world datasets (Section~\ref{sec:corr-analysis}),
a complexity and speedup measurement on the synthetic
and real-world datasets, respectively (Section~\ref{sec:complexity}),
and a qualitative neighbourhood analysis on a small illustrative graph
(Section~\ref{sec:neighbourhood}).

\subsubsection{Encoding Methods Under Comparison}
\label{sec:enc-methods}

Three variants of the $\GTS$ encoding
(Section~\ref{sec:g2s}) are evaluated,
ordered by decreasing computational cost:

\begin{description}[nosep, style=unboxed, leftmargin=0pt]
  \item[Canonical.]
    The exhaustive-backtracking procedure of
    Definition~\ref{def:canonical},
    returning the lexicographically minimal shortest string $\wstar_G$.
  \item[Greedy-min.]
    The greedy $\GTS$ algorithm executed from every starting node
    $v_0 \in V(G)$;
    the shortest string across all runs is selected.
  \item[Greedy-rnd($v_0$).]
    A single greedy $\GTS$ run from a uniformly random starting node.
\end{description}

\subsubsection{Distance Computation}

Exact graph edit distance
(GED; Definition~\ref{def:ged})
serves as the ground-truth structural dissimilarity measure.
The GED computation procedure for each dataset
is described in Section~\ref{sec:datasets}.

For each encoding method, all-pairs Levenshtein distance matrices
(Definition~\ref{def:levenshtein}) are computed
from the resulting \IsalGraph{} strings.

\subsubsection{Correlation Analysis}
\label{sec:corr-analysis}

The agreement between the Levenshtein and GED distance matrices
is quantified over all valid upper-triangular pairs $(i, j)$
with $i < j$,
$\mathrm{GED}(G_i, G_j) > 0$,
and $d_{\mathrm{Lev}}(w_i, w_j) > 0$.
Two statistics are reported per dataset and encoding method:

\begin{itemize}[nosep]
  \item Spearman's rank correlation coefficient $\rho$,
        measuring monotonic association between the two distance measures.
  \item The ordinary least-squares (OLS) regression slope $\beta$
        of Levenshtein distance on GED,
        where $\beta = 1$ indicates equal scaling and $\beta < 1$ indicates
        that Levenshtein distances grow more slowly than GED.
\end{itemize}

The $p$-values reported in Table~\ref{tab:performance-summary}
are obtained from SciPy's implementation of the Spearman test,
which uses the asymptotic $t$-distribution approximation
$t = \rho\sqrt{(n-2)/(1-\rho^2)}$ with $n - 2$ degrees of freedom,
where $n$ is the number of valid pairs.
Given that all five datasets yield $n > 1{,}600$ pairs,
the asymptotic approximation is well justified.
Statistical significance is assessed at $\alpha = 0.001$.

\subsubsection{Complexity and Speedup Measurement}
\label{sec:complexity}

Encoding time is measured on the synthetic graph families
described in Section~\ref{sec:datasets}.
Each encoding is repeated 25 times per graph instance,
and the median CPU time is retained to reduce the effect
of system scheduling variance.
The aggregate time at each node count $n$ is the median
across instances and families, with the interquartile range
as a dispersion measure.
Scaling exponents $\alpha$ are estimated by fitting
$T(n) = c \cdot n^{\alpha}$ via OLS on log-transformed data
($\log T$ vs.\ $\log n$),
and goodness of fit is reported as $R^2$.

The computational speedup of the \IsalGraph{} pipeline
(encoding plus pairwise Levenshtein distance) over exact GED
is measured on the five real-world datasets.
Speedup ratios are computed per graph pair
and aggregated as the geometric mean,
stratified by graph size
($n = 3$ to $11$ nodes).

\subsubsection{Neighbourhood Topology}
\label{sec:neighbourhood}

As a qualitative illustration of the locality property
(Section~\ref{sec:topology}),
the neighbourhood structure of a representative small graph
is examined under both GED and Levenshtein distance.
The base graph is the house graph on 5 nodes and 6 edges.
All graphs at $\mathrm{GED} = 1$ from the base graph
(single edge edits) are enumerated,
and their Levenshtein distances to the base encoding are computed.
Conversely, all strings at Levenshtein distance~$1$ from
the base encoding are generated
(single character substitutions, insertions, and deletions),
decoded via $\STG$,
and their GED to the base graph is computed.

\subsection{Implementation}
\label{sec:implementation}

The \IsalGraph{} core is implemented in Python
with no external dependencies.
Adapters for NetworkX~\citep{hagberg2008networkx},
igraph~\citep{csardi2006igraph},
and PyTorch Geometric~\citep{fey2019pyg}
provide interoperability with standard graph libraries.
Timing measurements use \texttt{time.process\_time()}
to record CPU time exclusive of I/O and system scheduling.
All experiments were executed on the Picasso supercomputer
at the Supercomputing and Bioinformatics Centre (SCBI)
of the University of M\'{a}laga.
A fixed random seed of~$42$ is used throughout
for reproducibility.

\section{Results}

We evaluate \IsalGraph{} along three axes:
agreement between Levenshtein distance and GED
(Section~\ref{sec:res-correlation}), and
empirical time complexity of the encoding algorithms
(Section~\ref{sec:res-complexity}).
A qualitative neighbourhood analysis
completes the evaluation (Section~\ref{sec:res-neighbourhood}).

\subsection{Correlation with Graph Edit Distance}
\label{sec:res-correlation}

Table~\ref{tab:performance-summary} reports the Spearman rank
correlation coefficient $\rho$ between GED and Levenshtein distance
for each dataset and encoding method.
All fifteen $\rho$ values are statistically significant
at $\alpha = 0.001$.

On the three IAM Letter subsets,
which contain sparse graphs ($\bar{m} \leq 4.56$),
the canonical encoding attains strong monotonic agreement
with GED:
$\rho = 0.934$ on \textsc{Low},
$0.876$ on \textsc{Med},
and $0.682$ on \textsc{High}.
Greedy-min trails canonical by modest margins
($\Delta\rho = 0.027$, $0.014$, and $0.057$, respectively),
while Greedy-rnd($v_0$) incurs larger losses,
reaching $\Delta\rho = 0.228$ below canonical on \textsc{Low}.

On the denser LINUX and AIDS datasets
($\bar{m} = 8.35$ and $10.70$),
correlation drops markedly.
The best method on LINUX is Greedy-min ($\rho = 0.445$),
the only dataset where it surpasses canonical ($\rho = 0.433$),
by a margin of $0.012$.
On AIDS, canonical leads with $\rho = 0.349$.
The small difference on LINUX may reflect
the limited number of valid pairs ($n = 1{,}685$)
relative to the other datasets ($n \geq 131{,}148$),
which amplifies the effect of individual outlier pairs
on the rank correlation statistic.

\begin{figure}[htp]
    \centering
    \includegraphics[width=\linewidth]{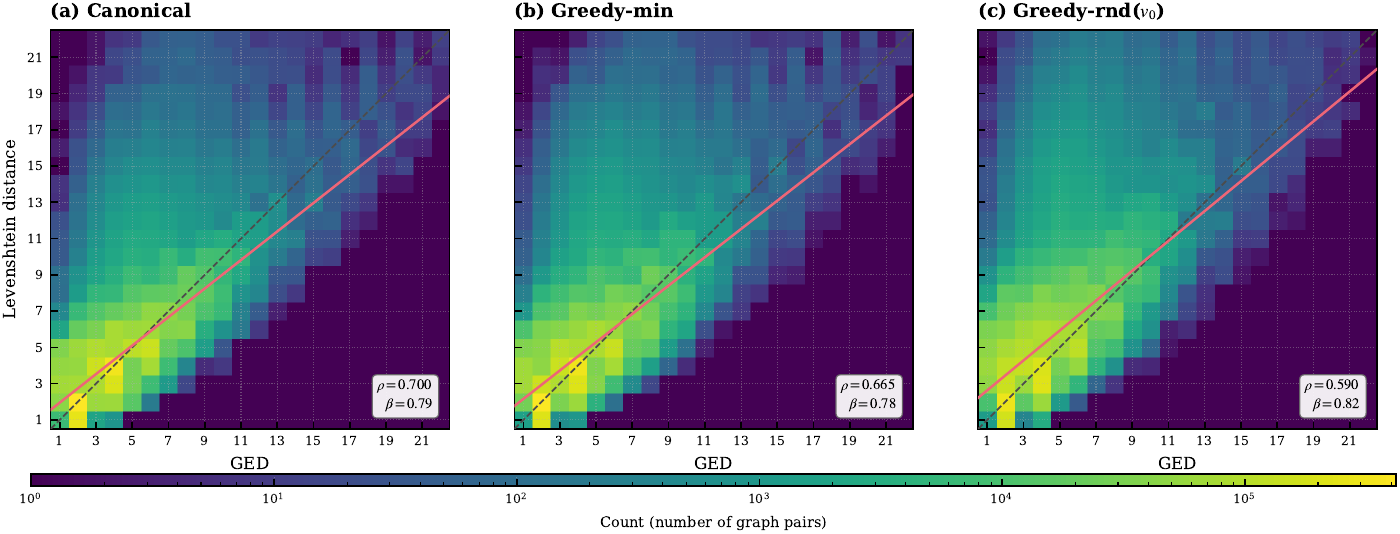}
    \caption{Aggregated correlation between graph edit distance (GED) and Levenshtein distance across all five benchmark datasets. Each cell at integer coordinates $(i, j)$ shows the count of graph pairs with $\text{GED} = i$ and $\text{Lev} = j$ (log scale; light = few pairs, dark = many pairs); white cells contain no observed pairs. Dashed grey line: identity ($\text{Lev} = \text{GED}$). Solid red line: ordinary least-squares (OLS) regression. (a) Canonical encoding ($n = 3,424,764$ pairs, $\rho = 0.700$, $\beta = 0.79$). (b) Greedy-min encoding ($n = 3,424,764$ pairs, $\rho = 0.665$, $\beta = 0.78$). (c) Greedy-rnd($v_0$) encoding ($n = 3,424,764$ pairs, $\rho = 0.590$, $\beta = 0.82$). Reported statistics: $\rho$ denotes Spearman's rank correlation coefficient, measuring monotonic association between the two distance measures. $\beta$ denotes the OLS regression slope; $\beta = 1$ would indicate that Levenshtein and GED operate on the same scale, while $\beta < 1$ indicates that Levenshtein distances grow more slowly than GED.}
    \label{fig:heatmap-correlation-ged-lev}
\end{figure}

\begin{table*}[hbp]
\centering
\caption{Dataset properties and Spearman $\rho$ correlation between GED and IsalGraph Levenshtein distance across encoding methods. $\bar{m}$: mean edges per graph (complexity proxy). Spearman-$\rho$ difference between best method per dataset is showcased. Best $\rho$ per dataset in \textbf{bold}.}
\label{tab:performance-summary}
\small
\begin{tabular}{clccccc}
\toprule
 &  & \textbf{IAM LOW} & \textbf{IAM MED} & \textbf{IAM HIGH} & \textbf{LINUX} & \textbf{AIDS} \\
\midrule
\multirow{3}{*}{\rotatebox[origin=c]{90}{\small Prop.}} & $N$ & 1{,}180 & 1{,}253 & 2{,}059 & 89 & 769 \\
 & Pairs & 695{,}610 & 784{,}378 & 2{,}118{,}711 & 1{,}685 & 131{,}148 \\
 & $\bar{m}$ & 3.07 & 3.17 & 4.56 & 8.35 & 10.70 \\
\midrule
\multirow{3}{*}{\rotatebox[origin=c]{90}{\small Spear. $\rho$}} & Canonical & \textbf{0.934}$^{***}$ & \textbf{0.876}$^{***}$ & \textbf{0.682}$^{***}$ & 0.433$^{***}$ (-0.012) & \textbf{0.349}$^{***}$ \\
 & Greedy-Min & 0.908$^{***}$ (-0.027) & 0.862$^{***}$ (-0.014) & 0.625$^{***}$ (-0.057) & \textbf{0.445}$^{***}$ & 0.304$^{***}$ (-0.045) \\
 & Greedy-rnd($v_0$) & 0.706$^{***}$ (-0.228) & 0.682$^{***}$ (-0.195) & 0.577$^{***}$ (-0.105) & 0.301$^{***}$ (-0.144) & 0.251$^{***}$ (-0.098) \\
\bottomrule
\end{tabular}
\vspace{1mm}\par\footnotesize $^{***}p<0.001$, $^{**}p<0.01$, $^{*}p<0.05$. $\bar{m}$ increases monotonically as $\rho$ degrades across datasets.
\end{table*}

Figure~\ref{fig:heatmap-correlation-ged-lev} displays
the aggregated joint distribution of GED and Levenshtein distance
over all $3{,}424{,}764$ valid pairs from the five datasets.
The concentration of mass near the identity line
confirms that the two measures are broadly co-monotonic,
though the spread increases at higher GED values.
The OLS regression slopes
$\beta = 0.79$ (Canonical), $0.78$ (Greedy-min),
and $0.82$ (Greedy-rnd)
lie consistently below unity,
indicating that Levenshtein distances grow more slowly than GED.
This compression stems from the bounded instruction alphabet
($|\Sig| = 9$):
structurally distant graphs can still share
long common subsequences in their encoding strings,
attenuating the measured string dissimilarity.

A monotonic relationship between graph density and
correlation strength is apparent across all five datasets.
As the mean edge count $\bar{m}$ increases
from $3.07$ (IAM \textsc{Low})
to $10.70$ (AIDS),
$\rho$ decreases for every encoding method
(Table~\ref{tab:performance-summary}).
The steepest drop occurs between
IAM \textsc{High} ($\bar{m} = 4.56$, canonical $\rho = 0.682$)
and LINUX ($\bar{m} = 8.35$, canonical $\rho = 0.433$),
where a near-doubling of mean edge count
coincides with a 37\% relative decline in $\rho$.
This degradation is consistent with the sequential nature
of the $\GTS$ traversal:
as edge density grows,
a single depth-first pass captures
a diminishing fraction of the graph's pairwise connectivity,
and the resulting string becomes
a coarser proxy for the full topology.

\subsection{Empirical Time Complexity}
\label{sec:res-complexity}

\begin{figure}[htp]
    \centering
    \includegraphics[width=0.7\linewidth]{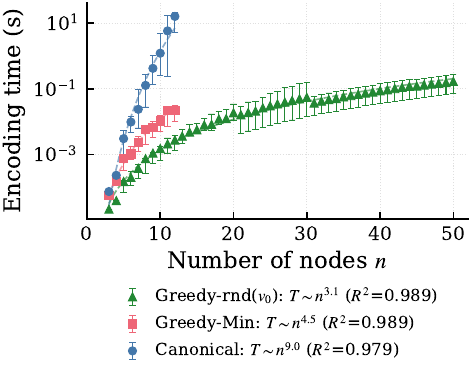}
    \caption{Empirical time complexity of IsalGraph encoding methods on random graphs (Barab\'asi--Albert $m \in \{1,2\}$ and Erd\H{o}s--R\'enyi $p \in \{0.3, 0.5\}$). Horizontal axis: number of nodes $n$; vertical axis: encoding time in seconds (log scale). Markers show the median across graph instances; error bars denote the interquartile range. Dashed lines are polynomial fits $T = c \cdot n^{\alpha}$ via OLS on log--log data. Greedy-rnd($v_0$): $\alpha = 3.1$, $R^2 = 0.989$. Greedy-Min: $\alpha = 4.5$, $R^2 = 0.989$. Canonical: $\alpha = 9.0$, $R^2 = 0.979$. Greedy methods exhibit polynomial scaling ($\alpha \approx 3$--$5$), while the canonical method scales super-polynomially ($\alpha \approx 9$) on random graphs and becomes infeasible beyond $n \approx 12$.}
    \label{fig:empirical-complexity}
\end{figure}

Figure~\ref{fig:empirical-complexity} shows the median encoding time
as a function of graph size $n$
for the three methods on synthetic random graphs
(Barab\'{a}si--Albert and Erd\H{o}s--R\'{e}nyi families).
Power-law fits $T(n) = c \cdot n^{\alpha}$
on log-transformed data yield exponents
$\alpha = 3.1$ for Greedy-rnd($v_0$),
$\alpha = 4.5$ for Greedy-min,
and $\alpha = 9.0$ for Canonical,
all with $R^2 \geq 0.979$.

The Greedy-rnd exponent $\alpha \approx 3$
is consistent with the cost of a single $\GTS$ traversal,
whose dominant operation is the neighbour-selection step
repeated at each of the $O(n)$ visited nodes.
Greedy-min iterates the greedy procedure over all $n$ starting nodes,
raising the empirical exponent to $\alpha \approx 4.5$;
the half-unit above $n^4$ reflects the variable string length
across starting nodes and the associated comparison cost.
Both greedy variants scale to graphs
with 50~nodes within the 600-second timeout.

The canonical method exhibits $\alpha = 9.0$,
a direct consequence of its exhaustive backtracking
over all starting nodes and all valid neighbour orderings.
At $n = 12$, the canonical encoding
already approaches the timeout threshold;
beyond this size, it is impractical
without further algorithmic refinement.
The high $R^2$ values confirm that the power-law model
captures the observed scaling within the tested range,
although the canonical method's true asymptotic complexity
is super-polynomial due to the combinatorial explosion
of traversal orderings.

\subsection{Neighbourhood Structure}
\label{sec:res-neighbourhood}

\begin{figure}[htp]
    \centering
    \includegraphics[width=\linewidth]{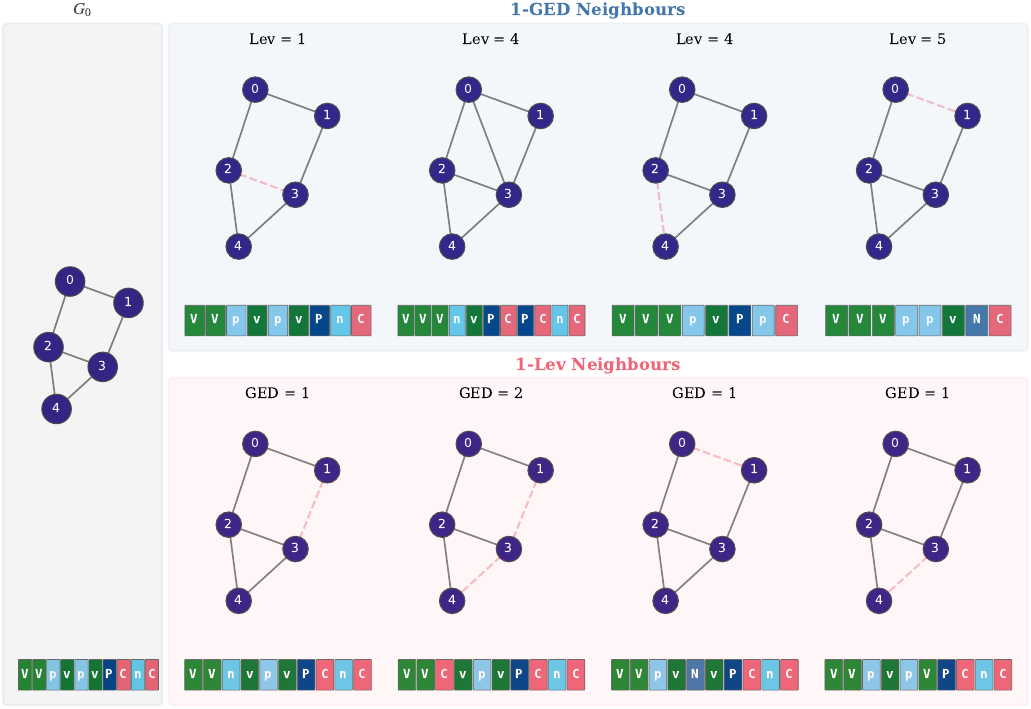}
    \caption{Neighbourhood topology of the house graph $G_0$ (5 nodes, 6 edges) under two distance metrics. Centre column: base graph $G_0$ with its canonical IsalGraph encoding (colour-coded by instruction type). Top rows: 4 representative 1-GED neighbours (single edge edit), with Levenshtein distances $\mathrm{Lev} \in [1,\, 5]$ to the encoding of $G_0$. Bottom rows: 4 representative 1-Levenshtein neighbours (single character substitution, insertion, or deletion in the instruction string), with GED values $\mathrm{GED} \in [1,\, 2]$. Dashed red edges indicate structural differences from $G_0$. Horizontal heatmaps below each graph render the IsalGraph instruction string with per-character colouring (alphabet $\Sigma = \{N,n,P,p,V,v,C,c,W\}$). The asymmetry between 1-GED and 1-Levenshtein neighbourhoods illustrates that graph-space proximity does not imply string-space proximity, and vice versa.}
    \label{fig:neighborhood-topology}
\end{figure}

Figure~\ref{fig:neighborhood-topology} illustrates the relationship
between graph-space and string-space proximity
on a concrete example,
complementing the aggregate correlation analysis above.
The base graph $G_0$ is the house graph (5~nodes, 6~edges),
and its canonical \IsalGraph{} encoding serves as the reference string.

The 1-GED neighbourhood of $G_0$
comprises 10~non-isomorphic graphs
obtained by a single edge insertion or deletion
that preserves connectivity
(6~deletions and 4~insertions).
Their Levenshtein distances to the encoding of $G_0$
range from 1 to~5:
a single structural edit can require
up to five character changes in the instruction string.
This spread arises because the canonical encoding
selects the globally optimal traversal order;
modifying one edge may shift this optimum entirely,
producing a substantially different string
even though the underlying graph changed minimally.

In the reverse direction,
the 1-Levenshtein neighbourhood---strings
differing from the base encoding
by a single substitution, insertion, or deletion---yields
graphs with $\mathrm{GED} \in \{1, 2\}$ to $G_0$.
String-space proximity thus implies graph-space proximity:
small perturbations to the instruction string
produce small structural changes.
This directional tightness follows from the instruction semantics,
where each character corresponds to at most one
structural operation (node creation, edge insertion, or pointer movement),
bounding the topological effect of any single character change.

The asymmetry between the two neighbourhoods---tight
from string-space to graph-space,
loose from graph-space to string-space---is inherent
to any encoding in which multiple traversal orders
can represent the same graph.
It has a practical implication:
Levenshtein distance on \IsalGraph{} strings
is more likely to overestimate GED
(when a small structural change requires
a large string rearrangement)
than to underestimate it
(since each character change has bounded structural impact).
This conservative bias favours recall over precision:
Levenshtein-based retrieval is more likely
to return a slightly dissimilar graph
than to miss a genuinely similar one,
a property that is advantageous
in retrieval settings where recall is prioritised.

\section{Conclusion\label{sec:Conclusion}}

\paragraph{Summary of contributions.}
This paper has introduced \IsalGraph{}, a sequential instruction-based
representation of finite simple graphs.  The encoding is defined by a
nine-instruction virtual machine that manipulates a circular doubly-linked
list (CDLL) of graph-node references via two traversal pointers, inserting
nodes and edges into a sparse graph as instructions are executed.  The
resulting representation has four properties that distinguish it from
existing graph encodings:
\begin{enumerate}[label=(\roman*)]
  \item \emph{Universal validity.}  Every string over the alphabet
        $\Sigma = \{N, n, P, p, V, v, C, c, W\}$ decodes to a valid
        finite simple graph.  There are no syntactically or semantically
        invalid strings, which eliminates the need for validity-checking
        decoders and simplifies the design of generative models.
  \item \emph{Reversibility.}  The greedy GraphToString ($\GTS$) algorithm
        encodes any connected graph $G$ into a string $w$ such that
        $\STG(w) \cong G$.  Round-trip correctness was confirmed at a
        100\% pass rate over 945 test instances spanning twelve graph
        families, with independent cross-validation via the VF2 isomorphism
        algorithm.
  \item \emph{Canonical completeness (conjectured).}  The canonical string
        $\wstar_G$, computed by exhaustive backtracking over all starting
        nodes and all valid neighbour orderings, is conjectured to be a
        complete graph invariant: $G \cong H \iff \wstar_G = \wstar_H$.
        This conjecture is supported by 100\% invariance and discrimination
        accuracy on 71 graph pairs across nine structural families including
        trees, cycles, complete graphs, stars, wheels, Barab\'asi--Albert
        graphs, and the Petersen graph.
  \item \emph{Metric locality.}  The Levenshtein distance between
        \IsalGraph{} strings correlates strongly with graph edit distance
        on real-world graph benchmarks, reaching Spearman $\rho = 0.934$
        on the sparse IAM Letter (LOW) dataset ($n = 695{,}610$ pairs,
        $p < 0.001$) and remaining significant across all five datasets
        tested, covering a range of structural densities from
        $\bar{m} = 3.07$ to $\bar{m} = 10.70$ mean edges per graph.
\end{enumerate}

\paragraph{Limitations.}
Three limitations warrant explicit acknowledgement.  First, the canonical
completeness conjecture remains unproven.  A formal proof would require
establishing rigorously that the exhaustive-backtracking algorithm is
label-blind, i.e.\ that its output depends solely on abstract adjacency
structure and not on the integer identifiers assigned to nodes.  Second,
the canonical encoding scales super-polynomially ($T \sim n^{9.0}$) and
is computationally infeasible for graphs with more than approximately 12
nodes on current hardware within a 600-second timeout.  Third, the $\GTS$
algorithm requires the input graph to be connected; for directed graphs, it
additionally requires all nodes to be reachable from the chosen starting
node via directed outgoing edges.  Graphs that do not satisfy these
conditions cannot be encoded without preprocessing.

\section*{Acknowledgment}
The authors thankfully acknowledge the computer resources (Picasso
Supercomputer), technical expertise, and assistance provided by the
SCBI (Supercomputing and Bioinformatics) center of the University
of M\'alaga.

\bibliographystyle{unsrtnat}
\bibliography{references}

\end{document}